\documentclass[twocolumn]{article}
\usepackage[utf8]{inputenc}

\usepackage{graphicx}

\usepackage{colortbl}
\usepackage[numbers]{natbib}
\usepackage{comment}
\usepackage{balance}
\usepackage{enumitem}
\usepackage{makecell}
\usepackage{array}
\usepackage{dirtytalk}

\usepackage{subfig}
\usepackage{amssymb}
\usepackage{amsmath}

\usepackage[final]{microtype}

\usepackage{algorithm}
\usepackage{algpseudocode}
\usepackage{fancyhdr}
\usepackage{ragged2e}

\usepackage{pdfpages}

\newcolumntype{L}[1]{>{\raggedright\let\newline\\\arraybackslash\hspace{0pt}}m{#1}}
\newcolumntype{C}[1]{>{\centering\let\newline\\\arraybackslash\hspace{0pt}}m{#1}}
\newcolumntype{R}[1]{>{\raggedleft\let\newline\\\arraybackslash\hspace{0pt}}m{#1}}

\fancypagestyle{firstpage}
{
    \fancyhf{}

    \fancyhead[L]{\small \justifying \noindent \textbf{Cite as: Bignold, A., Cruz, F., Dazeley, R., Vamplew, P., Foale, C. (2022). Human Engagement Providing Evaluative and Informative Advice for Interactive Reinforcement Learning. Neural Computing and Applications.} 
    }

    \fancyfoot[L]{\small \justifying \noindent This preprint has not undergone peer review or any post-submission improvements or corrections. The Version of Record of this article is published in Neural Computing and Applications, and is available online at https://doi.org/10.1007/s00521-021-06850-6.
    }
    
}

\title{Human Engagement Providing Evaluative and Informative Advice for Interactive Reinforcement Learning}

\author{
Adam Bignold$^{1,\dagger}$ \and 
Francisco Cruz$^{2,3,\dagger}$ \and 
Richard Dazeley$^{2}$ \and 
Peter Vamplew$^{1}$ \and 
Cameron Foale$^{1}$} 

\date{\normalsize
$^{1}$ School of Engineering, IT and Physical Sciences, Federation University, Ballarat, Australia.\\
$^{2}$ School of Information Technology, Deakin University, Geelong.\\
$^{3}$ Escuela de Ingenier\'ia, Universidad Central de Chile, Santiago.\\
Corresponding e-mails: \{a.bignold, p.vamplew, c.foale\}@federation.edu.au,  \{francisco.cruz, richard.dazeley\}@deakin.edu.au\\
$^{\ddagger}$ Both authors contributed equally to this manuscript.
}

\begin{document}

\maketitle

\begin{abstract}
Interactive reinforcement learning proposes the use of externally-sourced information in order to speed up the learning process. 
When interacting with a learner agent, humans may provide either evaluative or informative advice. 
Prior research has focused on the effect of human-sourced advice by including real-time feedback on the interactive reinforcement learning process, specifically aiming to improve the learning speed of the agent, while minimising the time demands on the human. 
This work focuses on answering which of two approaches, evaluative or informative, is the preferred instructional approach for humans.
Moreover, this work presents an experimental setup for a human-trial designed to compare the methods people use to deliver advice in terms of human engagement.
The results obtained show that users giving informative advice to the learner agents provide more accurate advice, are willing to assist the learner agent for a longer time, and provide more advice per episode.
Additionally, self-evaluation from participants using the informative approach has indicated that the agent's ability to follow the advice is higher, and therefore, they feel their own advice to be of higher accuracy when compared to people providing evaluative advice.
\end{abstract}


\thispagestyle{firstpage}

\section{Introduction}

\begin{figure*}
    \centering
    \includegraphics[width=0.75\linewidth]{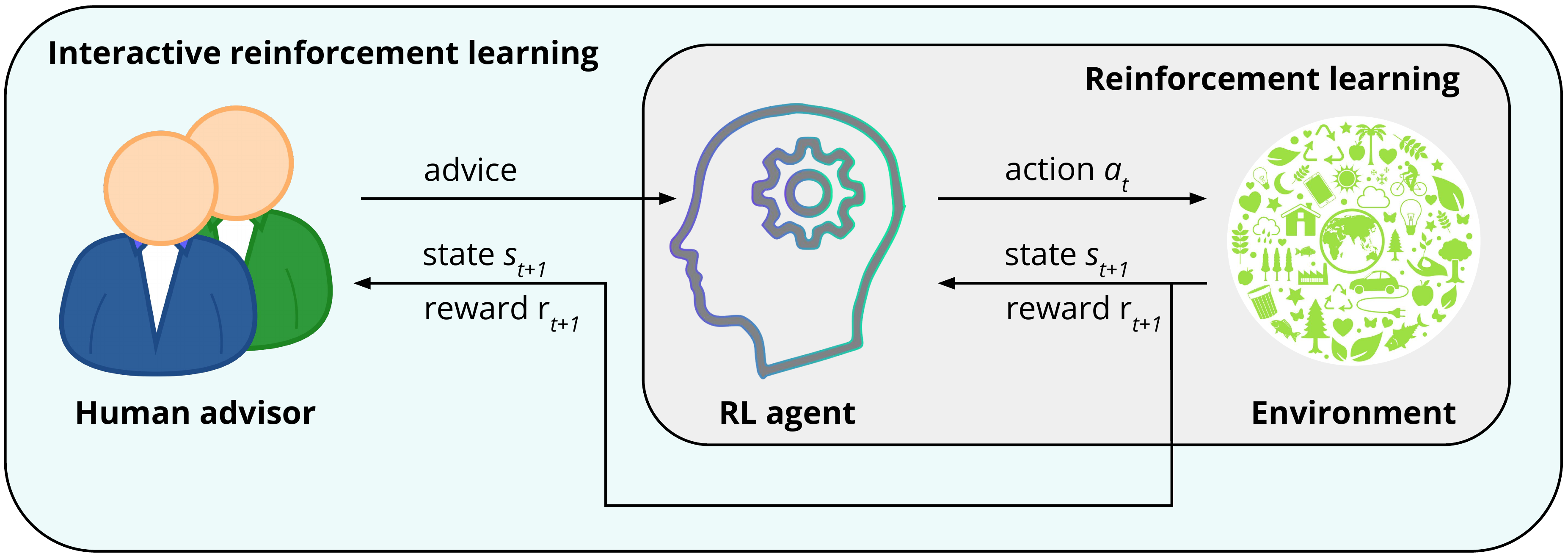}
    \caption{
    Interactive reinforcement learning.
    The autonomous agent performs an action $a_t$ from a state $s_t$ and the environment produces a response leading the agent to a new state $s_{t+1}$ and receiving a reward $r_{t+1}$.
    The interactive approach adds a human advisor for assistance.
    The advisor also observes the environment's response, and can provide either evaluative or informative advice to the learner agent.
    }
    \label{fig:HumanAdviceIntRL}
\end{figure*}

Reinforcement Learning (RL) aims at the creation of agents and systems that are capable of functioning in real-world environments~\cite{sutton2018reinforcement}. 
A common RL task involves decision-making and control, which given some information about the current state of the environment, must determine the best action to take in order to maximise long-term success. 
In this regard, RL allows improving the decision-making process while operating, to learn without supervision, and adapt to changing circumstances~\cite{rlsurvey}. 
In classical, autonomous RL~\cite{sutton2018reinforcement} the agent interacts with its environment learning by trial-and-error. 
The agent explores the environment and learns solely from the rewards it receives (see grey box within Figure~\ref{fig:HumanAdviceIntRL}). 
RL has shown success in different domains such as management~\cite{giannoccaro2002inventory, lepenioti2021human}, chemical processes~\cite{machalek2021novel, cruz2007indirect},  robot scenarios~\cite{robocup, churamani2020icub, lee2021pebble}, and game environments~\cite{tdgammon, barros2020moody}, among others.
However, RL has difficulties to learn in large state spaces. 
As environments become larger the agent's training time increases and finding a solution can become impractical~\cite{mankowitz2019challenges, cruz2018action}. 

Interactive Reinforcement Learning (IntRL) is an alternative to RL in which an advisor interacts with an RL agent in real-time~\cite{thomaz2005real}. 
The advisor can provide extra information to the agent regarding its behaviour or future actions it should perform. 
In this regard, the advice can be either evaluative or informative~\cite{li2019human}.
The former is an evaluation the advisor gives to the agent indicating how good or bad was the last action performed.
The latter is a suggestion given to the agent indicating what action to perform next from the current state.
From the agent's perspective these styles of interaction are commonly referred to as reward-shaping and policy-shaping respectively. 
Human advisors are usually used in IntRL since they achieve good performance in areas such as problem-solving, forward planning, and teaching.
Moreover, they have a large collection of knowledge and experiences to draw upon when encountering new environments and problems~\cite{brod2013influence}. 
IntRL utilise these skills of humans to assist the agent with its own learning and decision-making. 
This approach has been shown to considerably improve the agent's learning speed and can allow RL to scale to larger or more complex problems~\cite{scale}. 
Figure~\ref{fig:HumanAdviceIntRL} shows the IntRL approach with a human advisor included providing either evaluative or informative advice to the learner agent.

There are two major barriers to humans providing information to RL agents. 
The first is the time required by the human. 
In this regard, it is important that the mechanisms used to provide advice to the agent serve to reduce the number of interactions required~\cite{bignold2021evaluation}.
The second barrier is the skill needed by the human to provide the information. 
Humans usually need both programming skills and knowledge of the problem dynamics to encode information relevant to the agent's learning~\cite{arzate2020survey, bignold2021conceptual}. 
A principle of IntRL is that the method to provide information to the agent should be understandable and usable by people without programming skills or deep problem domain expertise~\cite{thomaz2005real, amershi2014power}. 
Therefore, the time required by a human advisor should remain as low as possible to reduce the burden on the human and methods for providing information to an agent should be accessible to users without programming or machine learning expertise.

Current studies on human engagement when teaching artificial agents have been mainly focused on assessing human commitment independent of the type of advice.
For instance, Thomaz and Breazel~\cite{thomaz2008teachable}  have shown that human tutors tend to have a positive bias when teaching agents, opting to reward rather than punish RL agents. 
It has been also shown that humans providing advice to a system over an extended period experienced frustration and boredom when receiving too many questions from the agent~\cite{cakmak2010optimality}. 
In this regard, studies suggest that participants do not like being prompted for input repeatedly, particularly when the input can be repetitive~\cite{bignold2021persistent}. 
Current IntRL systems require the users to provide advice on a state-by-state basis~\cite{moreira2020deep}, leaving current systems susceptible to the same issues of frustration and interruption as the active learning systems reported.

In this work, we aim to reduce the obligation of the human advisor while improving the learning speed of the agent. 
We address the question of which of the approaches, evaluative or informative, is the preferred instructional approach for humans.
To this aim, we carry out an analysis of human engagement with twenty participants with no prior knowledge of machine learning techniques in order to avoid any previous bias.
In our experiments, ten users give evaluative advice to the RL agent while ten users give informative advice in a simulated scenario.
From the performed interactions, we analyse the advice accuracy and the advice availability of each assistive approach. 
We also present an analysis of how evaluative advice may be affected by reward bias when teaching the RL agent.

Therefore, the distinction between advice delivery styles, i.e., evaluative or informative (also known as reward-shaping and policy-shaping respectively), and how humans engage and prefer to teach artificial agents is studied in this work.
While evaluative and informative approaches are about the method used to instruct the agent, reward-shaping and policy-shaping methods are about how the agent incorporates the provided advice, thus considering the agent's viewpoint. 
The main contributions of this paper can be summarised as follows:

\begin{itemize}
    \item Introduction of the concepts of evaluative and informative advice as methods to instruct autonomous agents.
    \item Analysis of human engagement when providing advice to a learning agent.
    \item Analysis of how evaluative advice is affected by reward bias.
    \item Comparison of evaluative and informative methods as preferred by human advisors.
    \item Evidence that users giving informative advice provide more accurate advice, are willing to assist the learner agent for a longer time, and provide more advice per episode.
\end{itemize}

\section{Reinforcement Learning and Interactive Human-sourced Advice}

Learning from the ground up can be a challenging task. 
While humans and artificial agents using RL are both capable of learning new tasks, it is evident that any extra information regarding the task can significantly reduce the learning time~\cite{cruz2018multi,sharma2007transfer,taylor2007transfer}. 
For humans, we can get advice from peers, teachers, the Internet, books, or videos, among other sources. 
By incorporating advice, humans can learn what the correct behaviour looks like, build upon existing knowledge, evaluate current behaviour, and ultimately reduce the amount of time spent performing the wrong actions~\cite{shin2020biased}. 
For artificial agents, the benefits of advice are the same. 
For instance, advice may be used to construct or supplement the reward function, resulting in an improved evaluation of the agent's actions or increased the utility of the reward function requiring fewer experiences to learn a behaviour~\cite{grzes2017reward, marom2018belief}. 
The advice can also be used to influence the agent's policy, either directly or through the action selection method, in order to reduce the search space~\cite{millan2020robust, millan2021robust}.

There are many possible information sources for agents to use. 
For instance, external information can come from databases~\cite{shah2016interactive}, labelled sets~\cite{deep1,deep2}, cases~\cite{kang1995multiple,compton1991ripple}, past experiences~\cite{taylor2009transfer}, other agents~\cite{multi1,multi2}, contextual perception~\cite{cruz2016learning}, and from humans~\cite{learningbydemonstration}. 
Human-supplied advice is contextually relevant information that comes from a human as a result of observation or awareness of the agent's current behaviour or goal. 
This information is commonly used to supplement, construct, or alter the RL process.  
Human-sourced advice can be more noisy, inaccurate, and inconsistent than other information sources. 
However, the critical benefit is that the advice is contextually relevant and can be applied to aid the agent in its current situation or goal. 


IntRL may use human-sourced advice~\cite{millan2019human} or simulated-users~\cite{ayala2019reinforcement} to directly interact with the agent while it is learning/operating~\cite{thomaz2005real}. 
The focus for IntRL is limited to the use of advice during the learning process, not before or after. 
This limitation requires interactive techniques to be easy for an agent to get information from, and for humans to add information to so that the learning process is not slowed down. 
This limitation also means that the agent or policy should not be reset when new information is provided, as that is conceptually similar to creating a new agent rather than interacting with an existing one. 
When humans interact with the agent, they may either provide additional rewards in response to the agent's performance~\cite{thomaz2007asymmetric} or recommend actions to the agent to guide the exploration process~\cite{moreira2020deep}. 


\subsection{Evaluative Advice}
\begin{figure}[ht]
    \centering
    \includegraphics[width=1\linewidth]{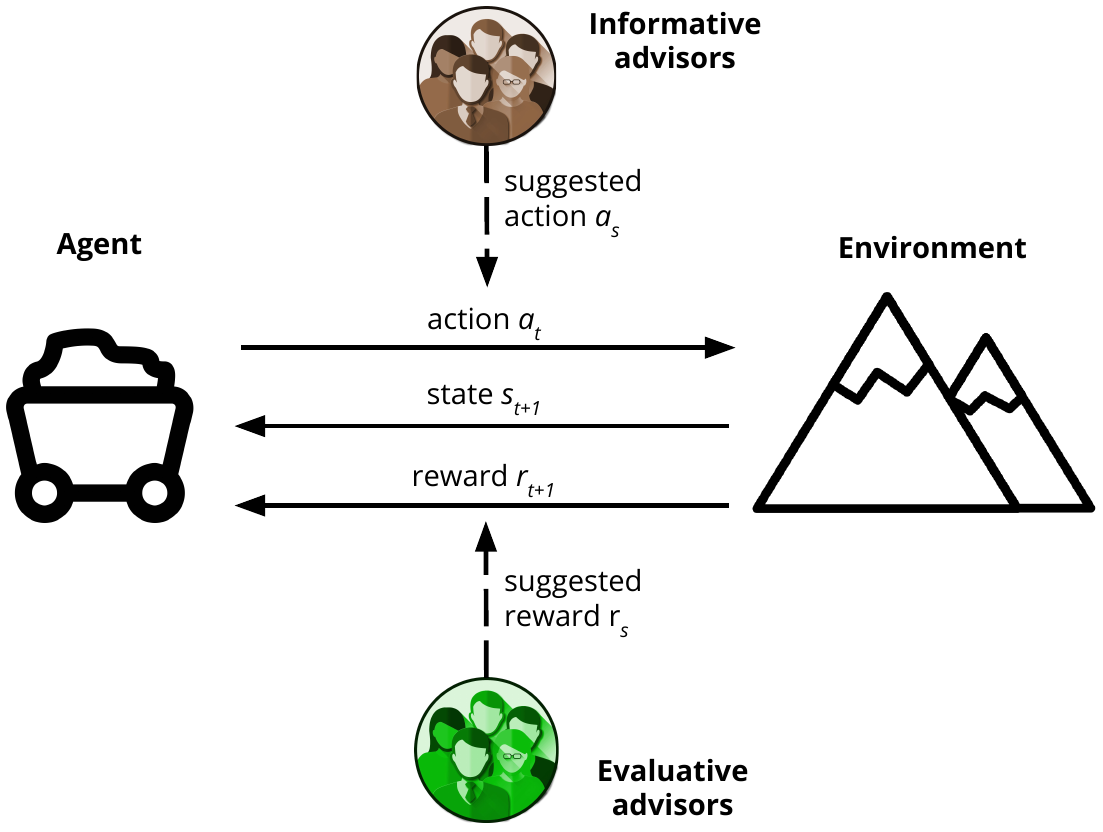}
    \caption{
    Interactive reinforcement learning feedback. 
    While the informative advisor may suggest an action to be performed by the agent, the evaluative advisor may suggest a reward to supplement the reward obtained from the environment.
    }
    \label{fig:InformativeEvaluativeIntRL}
\end{figure}

Evaluative advice is information that critiques current or past behaviour of an agent~\cite{ng1999policy, brys2014combining}.
Advice that supplements, improves, or creates a reward function is considered to be evaluative as it is a reaction to an agent’s behaviour rather than a direct influence on an agent's decision-making. 
The source of the advice is what separates evaluative advice from the reward function. 
A typical reward function is defined for an specific environment, whereas evaluative advice originates from an observer of the agent or other external sources~\cite{marthi2007automatic, bignold2021conceptual}. 
Figure~\ref{fig:InformativeEvaluativeIntRL} shows in green colour evaluative advisors supplementing the reward received from the environment.

Humans providing evaluative advice do not need to know the solution to a problem~\cite{rosman2014giving, huang2021gan}, it is enough for them to be able to assess the result of an action and then decide whether it was the correct action to take. 
For instance, in the training an agent manually via evaluative reinforcement (TAMER) framework~\cite{tamer, knox2009interactively}, a human user continually critiques the RL agent's actions. 
The human observes the agent, and in response to the agent's actions, provides a simple yes/no evaluation of its choice of action. 
This Boolean evaluation acts as an additional reward signal, supplementing the reward function from the environment. 
This bare minimum of human influence is enough to significantly decrease the time required by the agent to learn the required task~\cite{tamer}. 

Another example of evaluative advice is the convergent actor-critic by humans (COACH) approach~\cite{macglashan2017interactive}.
In this approach, a human trainer may give positive or negative feedback to a virtual dog learning to reach a goal position.
The human feedback was divided into punishment and reward and labelled with different levels as 'mild electroshock', 'bad dog', 'good dog', and 'treats'.
Using COACH, the agent was able to learn the task facing multiple feedback strategies.
Recently, this approach has been extended as Deep COACH~ \cite{arumugam2019deep} to represent the agent policy by deep neural networks.

\subsection{Informative Advice}
Informative advice is information that aids an agent in its decision-making~\cite{kessler2019active, paniz2020useful}.
Advice that recommends actions to take or avoid, suggests exploration strategies, provides information about the environment or proactively alters what action an agent may take is considered to be informative. 
Informative methods primarily focus on transferring information from the human and encoding it into the agent's policy, either directly, by altering the policy, or indirectly by influencing the agent's decision-making process~\cite{lin2020review}. 
Figure~\ref{fig:InformativeEvaluativeIntRL} shows in brown colour informative advisors suggesting an action to be taken.

Providing informative advice can be challenging for two reasons, the first of which is the human factor. 
Informative advice typically requires the human to know what the correct action is for a given state ahead of time. 
Not only does this require a greater understanding of the environment and the agent's position within it, but it also requires a more substantial commitment of time and effort to provide the advice. 
The time and effort required increases as the size of the environment, and the available actions increases~\cite{cruz2017agent}. 
The second reason utilising informative advice is challenging is that encoding information sourced from a human into a form an agent can understand can be a complicated process, as it is more informationally dense then evaluative advice~\cite{grizou2013robot}. 

For instance, an implementation of informative advice in IntRL is the ADVISE algorithm~\cite{griffith2013policy}. 
In ADVISE, a human observing an agent in operation can recommend actions to take at any given step, which the agent may choose to follow. 
This methodology allows the human to guide the agent through parts of the environment which they are familiar with. 
This can result in a significant improvement over existing IntRL methods and a reduced need for exploration.

Another example of informative advice was presented by Cruz et al.~\cite{cruz2018multi} in which a robot learned a cleaning task using human-provided interactive feedback. 
In this domestic scenario, seven actions could be advised to the agent using multi-modal audiovisual feedback. 
The provided advice was integrated into the learning process with an affordance-driven~\cite{cruz2016learning} IntRL approach. 
After experiments, the robot collected more and faster reward, tested against different minimal confident level thresholds and different levels of affordance availability.

\subsection{Evaluative versus Informative}
Evaluative advice has been more widely utilised in prior research as implementations are simpler to encode as the focus tends to be on the result of a decision rather than on what decision should be made~\cite{pilarski2012between}. 
This is due to it being easier to determine if an action was the correct or incorrect action to take once the result of the action is available. 
Most implementations of evaluative advice alter or supplement the reward function of the environment. 
Encoding information to alter the reward function is generally straightforward, as the primary focus is on whether to increase or decrease the reward given to the agent, as opposed to informative implementations that attempt to alter the decision-making policy~\cite{amir2016interactive}. 
Additionally, providing an evaluation requires less human effort than determining what information or action is relevant for a given state, as the information sought is typically a Boolean or a scalar measurement. 
Overall, evaluative advice is more direct to obtain, implement, and encode than the informative counterpart. 

Informative advice tends to be more informationally dense than evaluative advice. 
While this does make sourcing and encoding the information difficult, it does provide more benefit to the agent~\cite{pilarski2012between}.  
Evaluative advice only reinforces behaviour after that behaviour has been exhibited, whereas informative advice can promote or discourage behaviour before it is presented. 
Advice that recommends taking or avoiding actions will reduce the search space for the agent, resulting in improved learning time. 
The downside of this is that if the agent never performs actions that are preemptively discouraged, and the advice is not optimal, then the optimal policy may not be found~\cite{cruz2018improving}.

A direct comparison of the two styles is difficult as the implementations of human-sourced advice vary. 
Griffith et al.~\cite{griffith2013policy} compared the effects of informative versus evaluative advice on artificial agents using their informative algorithm ADVISE, against the evaluative algorithm TAMER. 
Both algorithms utilise IntRL agents and advice is given on a step by step basis. 
The ADVISE algorithm prompts the advisor for a recommended action which the agent can then follow, while TAMER prompts the advisor for a binary evaluation on the previously taken action. 
In the experiments, each agent is assisted by a simulated human, making the advice comparable. 

The ADVISE algorithm allows the advisor to recommend an action and therefore the number of bits of information provided is equal to $log_2(n_a)$ where $n_a$ is the number of possible actions (e.g., if there are eight possible actions $n_a = 8$, then each piece of informative advice provides three bits of information). 
In contrast, TAMER allows the human to provide a binary evaluation (i.e., correct/incorrect) which provides only a single bit of information.
Therefore the information gain from ADVISE is greater than TAMER and may bias the results. 
However, the experiments show that informative advice is more beneficial to the agent regardless of advice accuracy for the majority of cases. 
The use of a simulated human as an oracle in these experiments allowed for the provision of consistent advice that does not suffer from biases introduced by real humans. 
However if the behaviour of actual human advice-givers differs from that of the simulated human in terms of either accuracy and/or engagement, then the impact on agent behaviour may not reflect that observed in this study. 
Therefore it is important to develop an understanding of the properties of actual human advice.

\subsection{Human Engagement}
Studies on human engagement and teaching styles when engaging with interactive machine learning agents have previously been studied~\cite{amershi2014power,crowdsourcing}, however, they have been mainly focused on assessing human commitment independent of the type of advice. 
For instance, Amershi et al.~\cite{amershi2014power} presented a comprehensive study looking at the engagement between humans and interactive machine learning. 
The study included some case studies demonstrating the use of humans as information sources in machine learning. 
This work highlighted the need for increased understanding of how humans engage with machine learning algorithms, and what teaching styles the users preferred. 

A study by Thomaz and Breazeal~\cite{thomaz2008teachable}, later confirmed by Knox and Stone \cite{knox2012reinforcement}, found that human tutors tend to have a positive bias when teaching machines, opting to reward rather than punish RL agents. 
This bias leads to agents favouring the rewards provided by the human over the reward function of the environment. 
The positive bias was observed in humans providing evaluative advice, as it tends to be provided as a reward \cite{thomaz2008teachable}. 
Due to its characteristics, no such bias has been tested for or observed yet in informative-assisted agents. 
Knox and Stone~\cite{knox2013learning} later mitigated the consequence of the positive bias in RL agents by developing an agent that valued human-reward gained in the long term rather than the short term.

Another study performed by Cakmak and Thomaz~\cite{cakmak2010optimality} investigated the strategy of teachers when tutoring machine learning agents. 
The study found that humans providing advice to a system over an extended period experienced frustration and boredom when bombarded with questions from the agent. 
The stream of questions to the teachers caused some participants to “turn their brain off” or “lose track of what they were teaching” according to self-reports~\cite{cakmak2010designing}. 
Similar results were obtained using a movie recommendation system developed for Netflix, where participants were repeatedly asked to state if the system was right or wrong~\cite{guillory2011simultaneous,guillory2011online}. 

The previous studies suggest that participants do not like being prompted for input repeatedly, particularly when the input can be repetitive. 
Current IntRL systems do not prompt the user for information, instead, allowing the advisor to step in whenever they wish. 
Nevertheless, input into these systems is repetitive and requires the users to provide advice on a state-by-state basis \cite{moreira2020deep}, leaving current systems susceptible to the same issues of frustration and interruption as the active learning systems reported. 
Regardless, it is still not clear whether these issues will be translated into the IntRL scenarios.
Therefore, the remainder of this paper reports details and results of an experiment carried out to establish the characteristics of advice provided by humans interacting with an IntRL agent, and to assess whether these properties alter depending on whether evaluative or informative advice is being provided.

\section{Experimental Methodology}
In this section, we describe the IntRL methodology used during the experiments and frame the approach within an assisted RL framework.
Moreover, we outline the method to collect human advice including participants' characteristics, induction process, experiment details, and after-experience questionnaire.

\subsection{Interactive Reinforcement Learning Methodology}
Assisted reinforcement learning (ARL)~\cite{bignold2021conceptual} is a general framework proposed to incorporate external information into traditional RL.
The framework uses a conceptual taxonomy including processing components and communications links to describe transmission, modification, and modality of the sourced information.
The processing components comprise information source, advice interpretation, external model, and assisted agent, whereas the communications links are temporality, advice structure, and agent modification.
ARL agents aim to gather as much information from an external source as possible, as this can lead to improved performance within the environment. 
A concrete example of an ARL agent is an IntRL agent. 
As previously mentioned, an IntRL agent can be advised with externally-sourced information to support the learning process at any time of the training.

\begin{figure}[ht]
\centering
\includegraphics[width=1\linewidth]{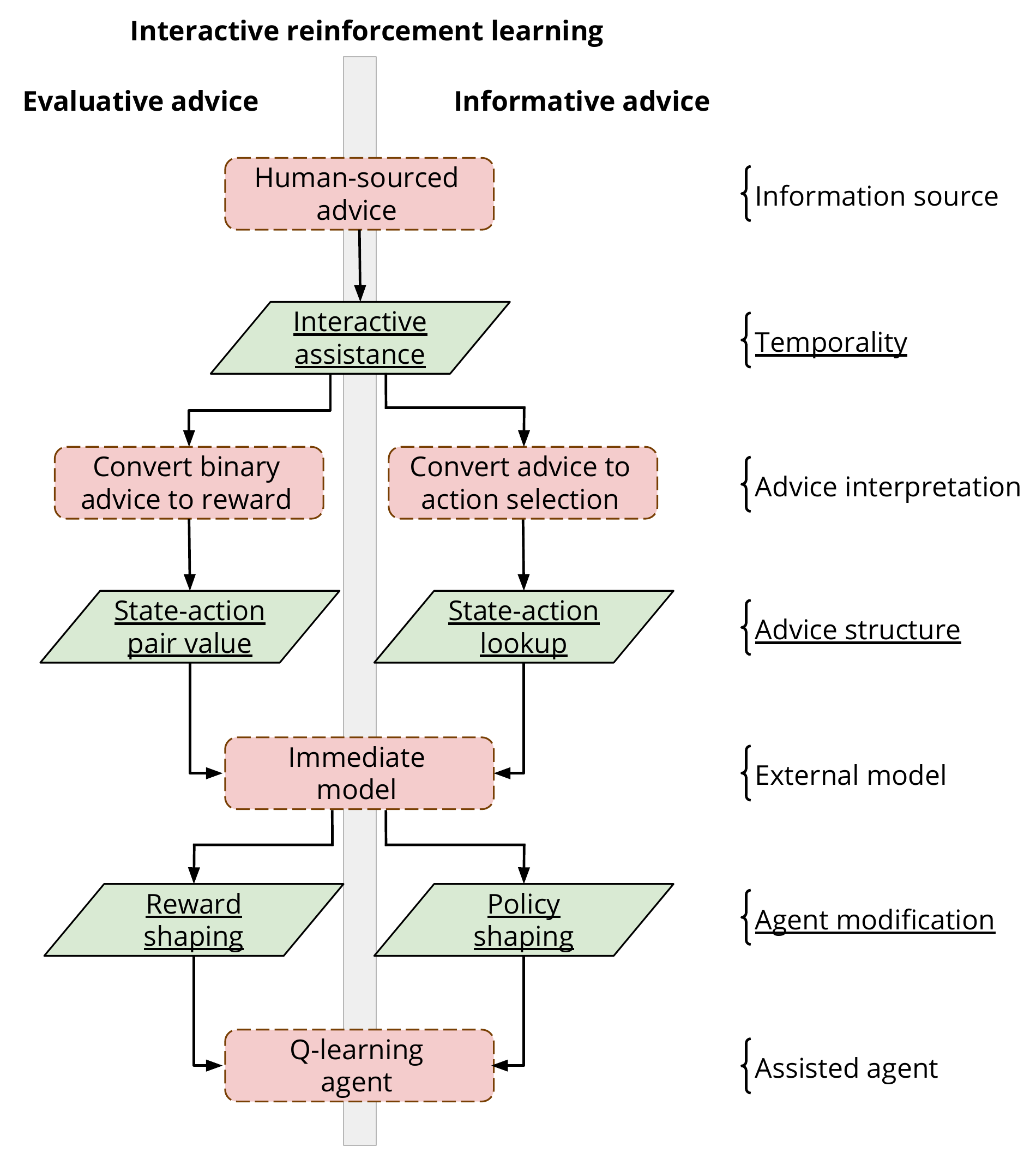}
\caption{
Interactive reinforcement learning method used to compare human engagement in evaluative and informative advice.
The method is presented using the assisted reinforcement learning taxonomy~\cite{bignold2021conceptual}, defining processing components (dotted red squares) and communication links (underlined green parallelograms) for each advice delivery style.}
\label{fig:EvaluativeInformativeIntRL}
\end{figure}


In this work, two different learner agents attempt to solve the Mountain Car problem~\cite{sutton2018reinforcement} using IntRL (more details about the experimental problem are given in the next section), the first agent accepts evaluative advice and the other receives informative advice.
Figure~\ref{fig:EvaluativeInformativeIntRL} shows the IntRL approach framed within the ARL framework~\cite{bignold2021conceptual} using both evaluative and informative advice.
The figure shows the processing components using dotted red squares and the communication links using green parallelograms with underlined text.
Using the ARL taxonomy, there are some common processing components and communication links that are adopted similarly by both approaches.
The common elements are information source, temporality, external model, and the assisted agent, which are adopted by the ARL framework as human-sourced advice, interactive assistance, an immediate model, and a Q-learning agent.
All the other processing components and communication links differ to each other for evaluative and informative advice.
For the evaluative approach, advice interpretation, advice structure, and agent modification are adopted by the ARL framework as binary advice to reward conversion, state-action pair value, and reward-shaping respectively.
For the informative approach, they are adopted as advice to action selection conversion, state-action lookup, and policy-shaping respectively. 
As this approach relies on human trainers as an external information source, the higher the people engagement, the higher the opportunity to transfer knowledge to the agent. 
The accuracy of the advice and information gain as a result of the advice provided is also important, as they contribute to the policy being learned by the agent~\cite{cruz2018improving}.

We aim to measure the human engagement, accuracy of advice, and the information gain for evaluative and informative advice for IntRL. 
To this aim, we perform experiments using two IntRL agents implemented with the temporal-difference learning method Q-learning. 
The performance of the agent, or its ability to solve the problem, is not the main focus of this paper. 
A comparison of evaluative and informative advice, in terms of the performance of the agents, has been investigated in a prior study~\cite{griffith2013policy}.

In the context of this work, human engagement is a measure of the number of interactions, the total time spent constructing interactions, and the distribution of interactions over the time the agent is operating. 
The observing human is given an opportunity to provide information once per step of the agent, and if the human does provide some advice during that step, then the interaction is recorded. 
However, a measure of the number of interactions is not sufficient, as the time and effort required to provide an interaction may differ between informative and evaluative advice methods. 
As a result, the interaction time is also recorded. 
Moreover, the accuracy of the information provided to the agent affects its performance within the environment~\cite{cruz2018improving}. 
In this regard, advice accuracy is a measure of how accurate the information provided by the human is, compared to the optimal action to take for each state the agent encounters. 
This can be calculated by comparing the advice provided by the human against the known optimal policy for this task.

\subsection{Human-sourced Advice}

During the experiments, twenty people participated, ten for each advice delivery style\footnote{Ethics approval letter provided by the Human Research Ethics Committee of \textit{Anonymous} University, research project number  \textit{Anonymous}.}
Each participant was able to communicate with an RL agent while observing its current state and performance. 
A participant interacting with the evaluative agent had the option of providing an agreement or disagreement (yes/no) to the agent's choice of action for the last time step. 
This binary evaluation was then used by the agent to supplement the reward it receives from the environment. 
A positive evaluation added $+1$ to the reward, while a negative evaluation subtracted $-1$ from the reward. 
Likewise, a participant interacting with the informative agent had the option of suggesting an action for the agent's next step, either left or right. 
If the agent was recommended an action then that action was taken, otherwise, the agent operates as a usual RL agent. 
Each participant, regardless of teaching style, had three possible options each step. 
For the evaluative advice participants, the options were: agree, disagree, or do nothing. 
Whereas, for the informative advice participants, the options were to recommend: left, right, or to do nothing. 

The participants chosen for the experiment had not had significant exposure to machine learning, and were not familiar with the Mountain Car environment. 
Before beginning the experiment, each participant was given a five-minute induction to the Mountain Car problem, and then asked to complete a short questionnaire. 
The induction introduces the aim of the agent, the dynamics of the environment, the action space, and most significantly, what the optimal solution to the problem is. 
The solution for the environment is described to the participant to give all participants an equal understanding and to reduce the time that they spend exploring the environment themselves so that they may focus on assisting the agent.

When the induction was complete, the participant was asked to complete a questionnaire. 
The full questionnaire consists of seven questions, the first two of which aim to assess the level of general knowledge about machine learning techniques and understanding of the Mountain Car problem of the participants.
After completing the first two questions, the participant is ready to begin the experiment.
The remaining five questions were answered after the subject had completed their interaction with the agent.

The participant was given 500ms to provide advice to the agent each step. 
To provide advice to the agent, the participant pressed one of two keys on the keyboard to indicate either approval/disapproval of the agent's last choice in action when using evaluative advice, or to recommend the left/right action for the agent to take next when using informative advice. 
Therefore, the input mechanism was dependent on the advice delivery style being tested. 
If the human provided advice within the 500ms window, an interaction had taken place and the time taken to create that interaction was recorded. 
If the human did not provide advice within the time window provided, then no interaction was recorded, and the agent operated as usual. 
Additionally, the human could change the duration of the time window by 25\% during the experiment by pressing the +/- keys. 
The experiments ran until the participant believed the agent had learned the correct behaviour, or until they tired of providing advice at which point the agent was terminated. 

After the participant had chosen to stop providing advice, they were asked to complete the remainder of the questionnaire. 
The remaining five questions aimed to assess understanding of the Mountain Car problem now that the participants have experienced the environment. 
It also aimed to capture their perception about their level of engagement, the accuracy of their advice, and the agent's understanding of the advice supplied.

\section{Interactive Reinforcement Learning Scenario}
In this section, we describe the key features of the experimental environment including the agent's representation, state and action representation, and reward function.
Furthermore, we complement the human-agent interactive methodology described in the previous section by indicating the script given to the participants.

\subsection{Features of the Environment}

\begin{figure}[ht]
    \centering
    \includegraphics[width=1\linewidth]{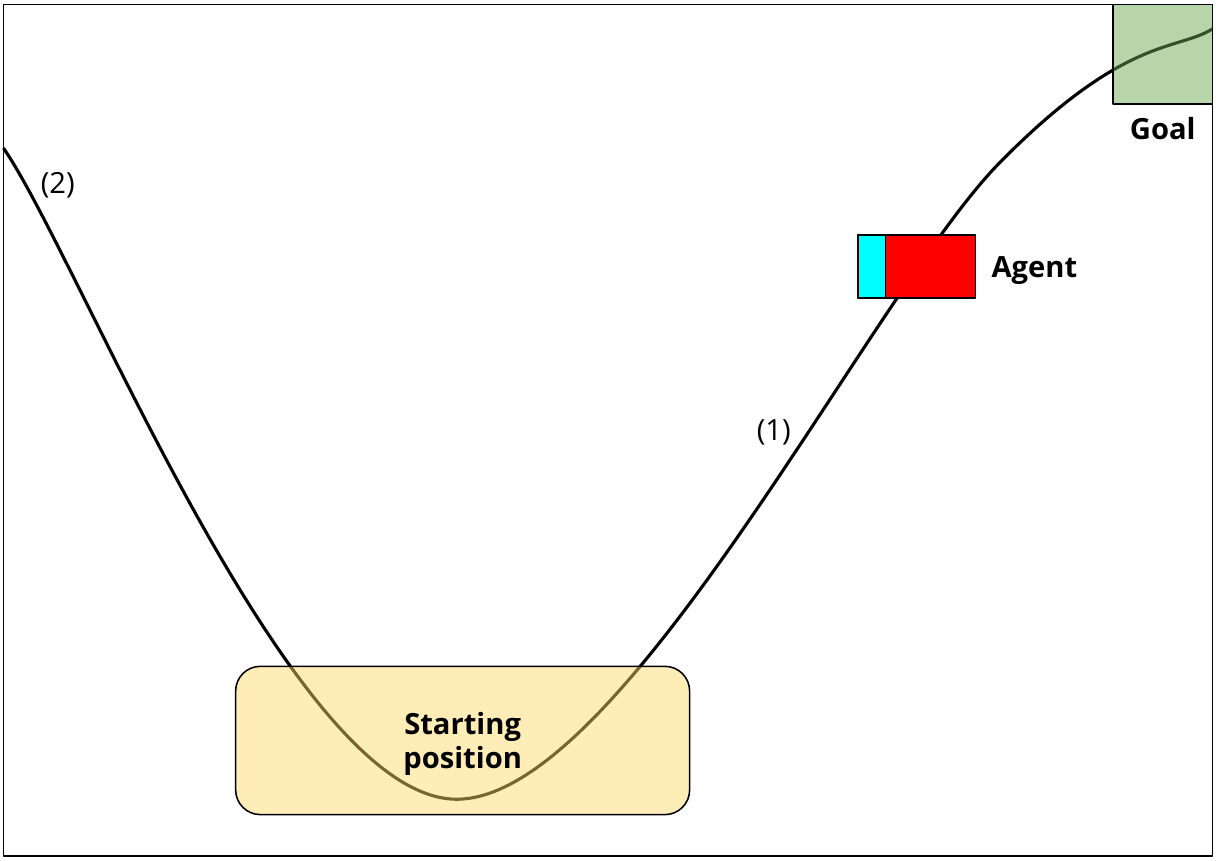}
    \caption{
    The Mountain Car environment. 
    The agent begins on the line at a random position within the yellow box and must travel to the green goal state. 
    To do so, the agent accelerates towards the first (1) key position until its velocity is reduced to zero by gravity. 
    At this point, the agent turns and accelerates towards the second (2) key position, again, until its velocity is reduced to zero. 
    Finally, the agent accelerates to the right again, building up velocity to reach the goal state.
    }
    \label{fig:MountainCarScenario}
\end{figure}

The Mountain Car environment is a standard continuous-state testing domain for RL~\cite{sutton2018reinforcement, moore1991variable}. 
In the environment, an underpowered car must drive from the bottom of a valley to the top of a steep hill. 
Since the gravity in the environment is stronger than the engine of the car, the car cannot drive straight up the side of the mountain. 
In order for the car to reach the top of the mountain, it must build up enough inertia and velocity. 
Figure~\ref{fig:MountainCarScenario} illustrates the mountain car environment and its features.

In our experiments, an RL agent controls the actions of the car. 
The car begins at a random position and with a low velocity somewhere within the starting position. 
In order to reach the goal position, the agent must build up enough momentum. 
To do so, the agent accelerates towards the goal until its velocity is reduced to zero by gravity. 
At this point, the agent turns and accelerates towards the other direction toward the highest possible position, again, until its velocity is reduced to zero. 
Finally, the agent accelerates down the hill again, building up velocity to reach the goal state. 
Should the agent not reach high enough up the mountain to reach the goal position, it should repeat the actions of accelerating in the opposite direction until a zero velocity is reached and turning around.

The key to the agent solving the Mountain Car problem is to increase its own velocity ($v$). 
The agent's mass ($m$), the magnitude of acceleration ($a$), and the force of gravity ($G$) are constant. 
As the agent's acceleration is lower than the gravity acting upon it, pulling the agent to the lowest point of the environment, the agent must accelerate at the correct moments, and in the correct direction, to increase its velocity. 
The optimal solution to the Mountain Car problem is to accelerate in the current direction of travel and take a random action when velocity is zero. 
An example of a rule formulation denoting this behaviour is shown in Eq. (\ref{eq:reward}).
The policy states the agent's next action ($A_t$) should be to accelerate right if its velocity is greater than 0, i.e. keep right movement, to accelerate left if its velocity is less than 0, i.e., keep left movement, and to take a random action if velocity is 0.

\begin{equation}
    A_t = \left\{
    \begin{array}{l r}
          +1 & v > 0\\
          -1 & v < 0\\
          \in\{-1, 1\} & v = 0\\
    \end{array}
    \right.
    \label{eq:reward}
\end{equation}

\begin{figure}[ht]
    \centering
    \includegraphics[width=1\linewidth]{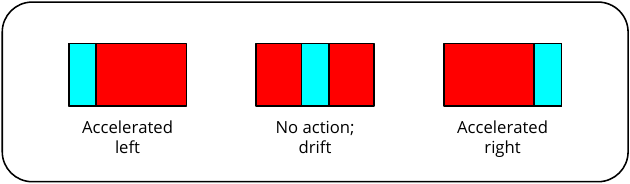}
    \caption{A graphical representation of the Mountain Car agent. 
    The entire rectangle (blue and red) represents the car. 
    The blue box indicates which action the agent has chosen to perform, either to accelerate left, to accelerate right, or not to accelerate at all and continue moving in its current direction of travel.}
    \label{fig:MountainCarActions}
\end{figure}

The agent controlling the car has three actions to choose from in any state: to accelerate left, to accelerate right, or not to accelerate at all.
The graphical representation of these possible actions is shown in Figure~\ref{fig:MountainCarActions}. 
At each step, the agent receives a reward of $-1$, and no reward when reaching the goal state. 
This reward encourages the agent to reach the goal in as few steps as possible to maximise the reward. 

The agent's state consists of two state variables, position and velocity, which are represented as real numbers. 
The position variable $p$ represents the agent's position within the environment, and ranges linearly from $-1.2$ to $0.6$, i.e. $p \in [-1.2, 0.6]$, with the lowest point of the environment residing at $-0.53$. 
The velocity of the agent $v$ has a range of $-0.07$ and $0.07$, i.e. $v \in [-0.07, 0.07]$. 
A velocity greater than zero indicates the agent is travelling to the right or increasing its position. 
If the agent collides with the edge of the environment on the left ($p=-1.2$) then the agent's velocity is set to 0. 

In this work, the RL agent utilises discrete state variables. 
Therefore, twenty bins for each state variable has been used, creating a total of 400 ($20\times20$) states. 
Of these 400 states, there are some that may never be visited by the RL agent, for example, it is impossible that the agent will be at top of the left mountain ($p=-1.2$) and have a high positive velocity ($v=0.07$).

\subsection{Interaction with Experiment's Participants}

As indicated in the previous section, twenty persons participated as trainers. 
The participants were university students with no experience in machine learning. 
Although we were aware that the number of participants was rather small, we were still able to draw significant conclusions for future experiments. 
During the experiments, the agents were given a low learning rate, manually tuned to extend the time which the agent would take to learn a suitable behaviour on its own. 
This was chosen so that the focus would be on the human's input rather than on the agent's capabilities. 
Both the evaluative and informative agents were given a learning rate of $\alpha=0.25$, a discounting factor of $\gamma=0.9$, and used an $\epsilon$-greedy action selection strategy with an epsilon of $\epsilon=0.05$. 

In order to avoid the participants get too familiar with the environment and biased the training, each person ran only one learning episode. 
The Mountain Car environment has been chosen since an optimal policy for the problem is known. 
To have an optimal policy for the environment allows the accuracy of the human-sourced information to be measured. 
Additionally, the Mountain Car problem has a low state and action space, allowing for the humans to observe the impact of their interactions relatively quickly, as the agent is likely to encounter the same state-action pairs frequently.

At the beginning of the experiments, the script given to the participants for describing the optimal solution is outlined below: 

\begin{quote}
``The car [agent] begins at the bottom of the valley, between two mountains. 
The car aims to drive to the top of the mountain on the right side. 
However, the car does not have the power to drive directly up the mountainside; instead, it needs to build up momentum. 
Momentum is gained by driving as high as possible on one side of the mountain, then turning around and accelerating in the opposite direction. 
When the car reaches the highest point it can on the opposite side, the process is repeated. 
Eventually, the car will gain enough speed to reach the top of the mountain.''
\end{quote}

\section{Results and Discussion}

In this section, we analyse the main results obtained from the experimental scenario. 
First, we present user's self-evaluations in terms of the level of task understanding, engagement with the interactive agent, self-reported accuracy, and ability to follow advice. 
Thereafter, we discuss the characteristics of the given advice, such as frequency, accuracy, and availability.

\subsection{User's Self-Evaluation}

As previously mentioned, before each participant began interacting with the agent, they were asked to answer two questions from the questionnaire. 
The purpose of the questionnaire is to assess the participants understanding of the problem environment and their interactions with the agent. 
The first question asked was whether the participant had previously been involved in a machine learning study.
None of the twenty participants reported being involved in a machine learning study previously.

\begin{figure}[ht]
    \centering
    \includegraphics[width=1\linewidth]{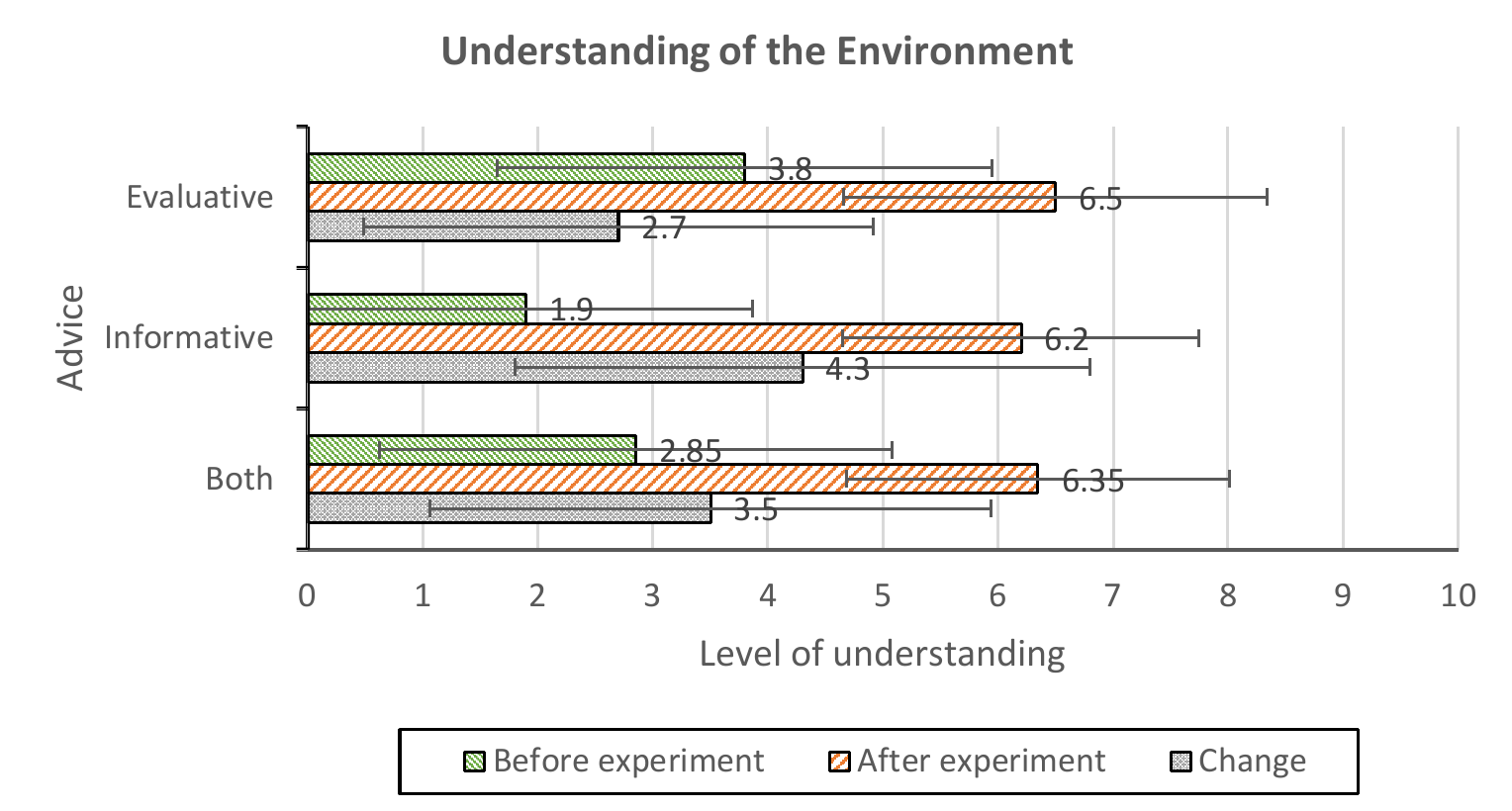}
    \caption{
    Participants' self-reported level of understanding of the solution and dynamics of the Mountain Car environment. 
    The standard deviation is shown over the bars for each approach and group.}
    
    \label{fig:SelfUnderstanding}
\end{figure}

\begin{figure}[ht]
    \centering
    \includegraphics[width=1\linewidth]{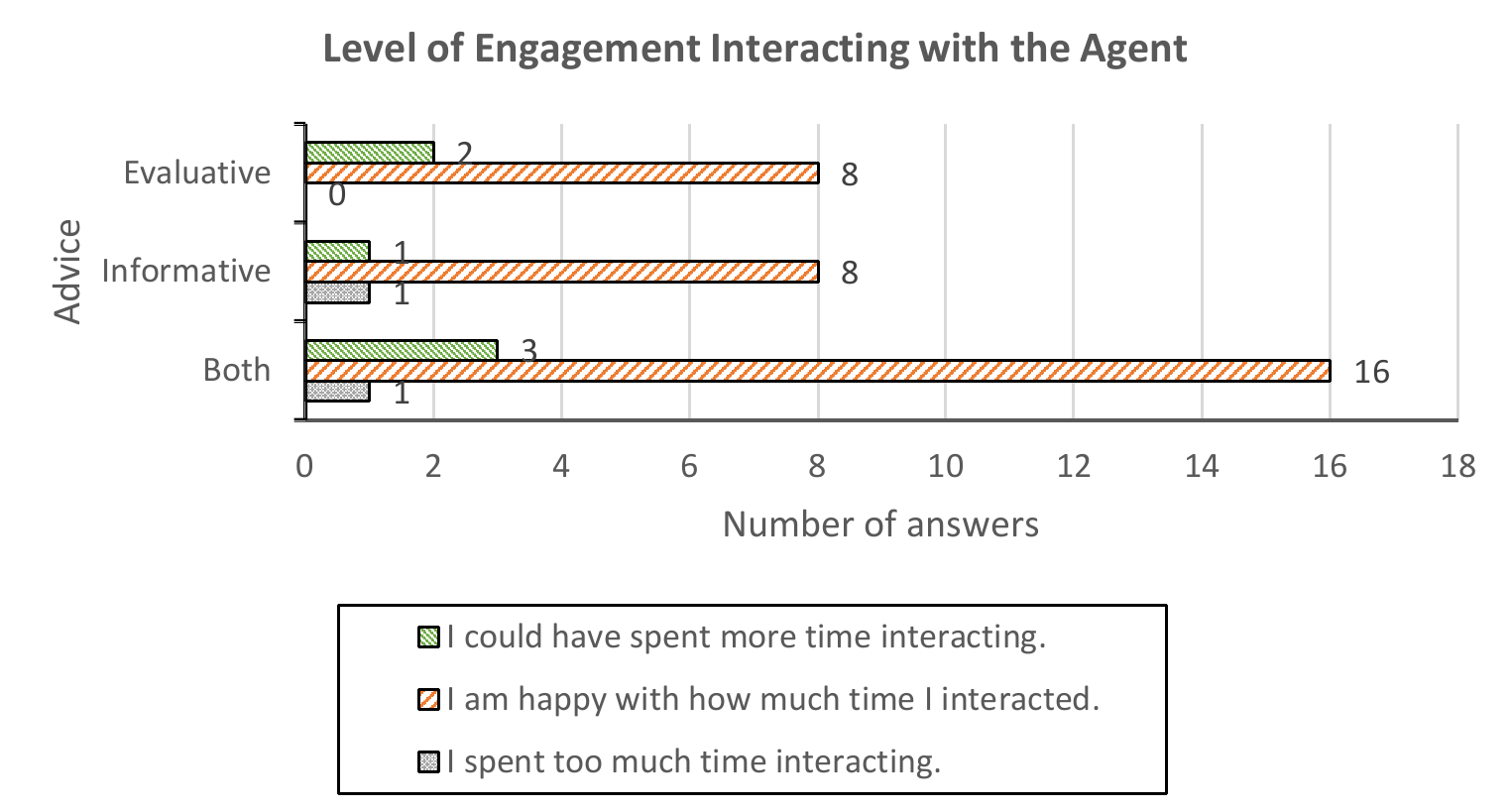}
    \caption{
    Participants' self-reported level of engagement with the agent. 
    Participants reported that they (a) could have spent more time with the agent, (b) were happy with how much time they provided, or (c) spent too much time with the agent. 
    }
    \label{fig:SelfEngagement}
\end{figure}

Participant were then provided with a brief explanation of the dynamics of the environment and what would be the optimal behaviour. 
Subsequently, before starting the experiment, they were then asked to rate their level of understanding of the environment on a scale of 0 to 10.
After interacting with the agent, the participants were asked the same question again. 
Figure~\ref{fig:SelfUnderstanding} shows the average self-reported level of understanding from the two groups of participants, i.e., evaluative and informative groups, and both before and after the experiments.

Interestingly, there is a small difference in the participant self-reported understanding of the environment before they begin interacting with the agent. 
The only difference in the explanation given to the two groups was the details on how they can interact with the agent. 
The participants giving evaluative advice were asked to rate the agent's choice of action as good or bad, while the participants giving informative advice were asked to recommend an action, either left or right. 
The difference in reported understanding before the experiment may indicate that evaluative advice delivery is easier to understand. 

Additionally, a change in the level of participants self-reported understanding is observed after the experiment. 
Although the informative group shows a greater change of understanding than the evaluative group after the experiment, this is due to the initial self-reported understanding. 
After assisting the agent, the two groups reported a greater understanding of the environment showing no significant difference between both of them. 

Moreover, after finishing the experiment, participants were also asked to report how they felt about their level of engagement with the agent.
They were given three different options to answer. 

\begin{enumerate}[label=(\alph*)]
\item I could have spent more time interacting with the agent.
\item I am happy with how much time I interacted with the agent.
\item I spent too much time interacting with the agent.
\end{enumerate}

Figure~\ref{fig:SelfEngagement} shows the participants' reported level of engagement with the agent indicating no significant difference between the two groups. 
In both cases, the majority of participants were content with the level of engagement they had with the agent.

\begin{figure}[ht]
    \centering
    \includegraphics[width=1\linewidth]{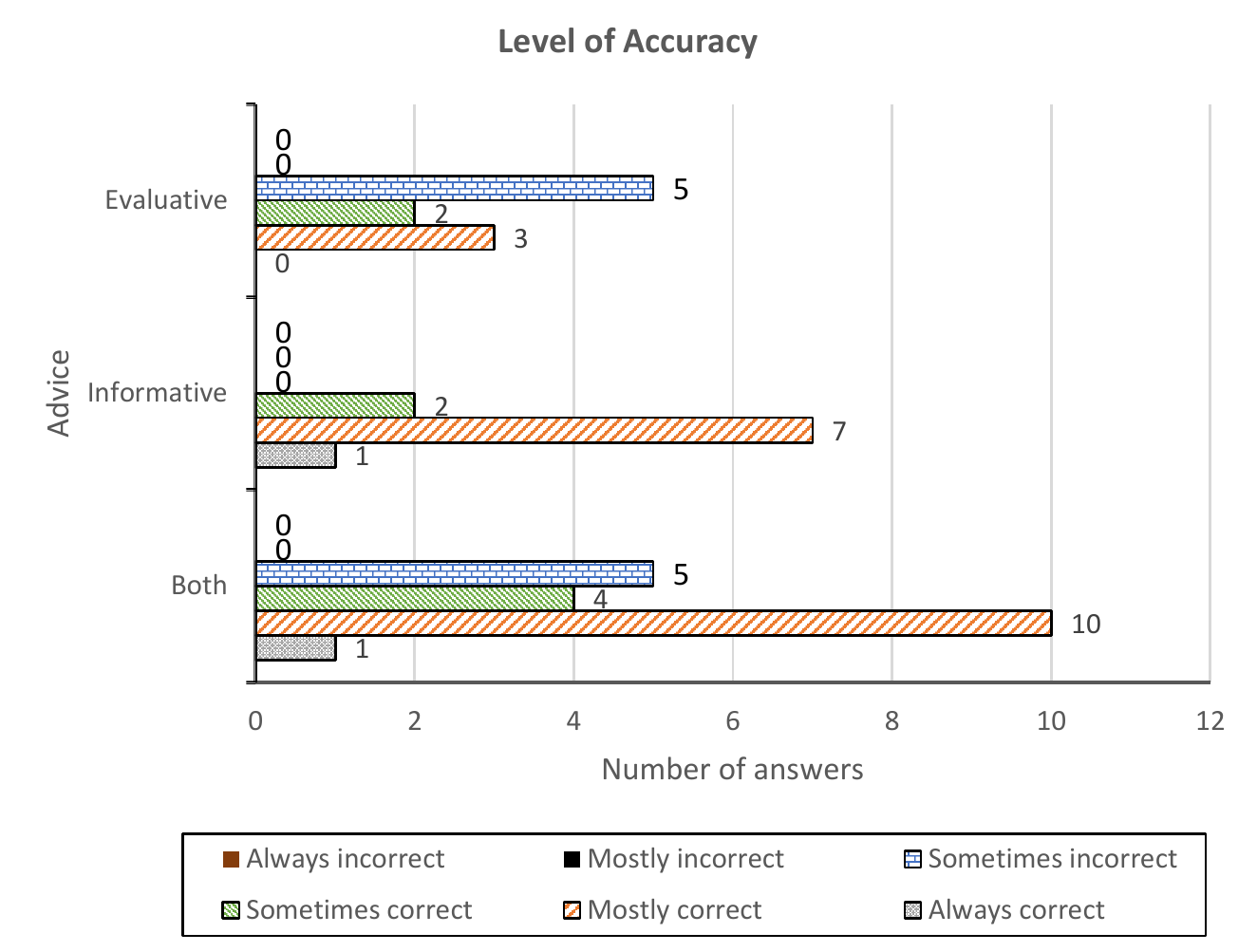}
    \caption{
    Participants' self-reported level of advice accuracy. 
    The informative group shows more confidence in the advice they give to the agent.
    }
    \label{fig:SelfAccuracy}
\end{figure}

The participants were asked to report what they thought their level of accuracy was throughout the experiment. 
Participants were given six different options to answer, ranging from always incorrect to always correct. 
Figure~\ref{fig:SelfAccuracy} shows the self-reported results. 
The results obtained indicate that participants in the informative group were more confident in the advice they provided to the agent. 

\begin{figure}[ht]
    \centering
    \includegraphics[width=1\linewidth]{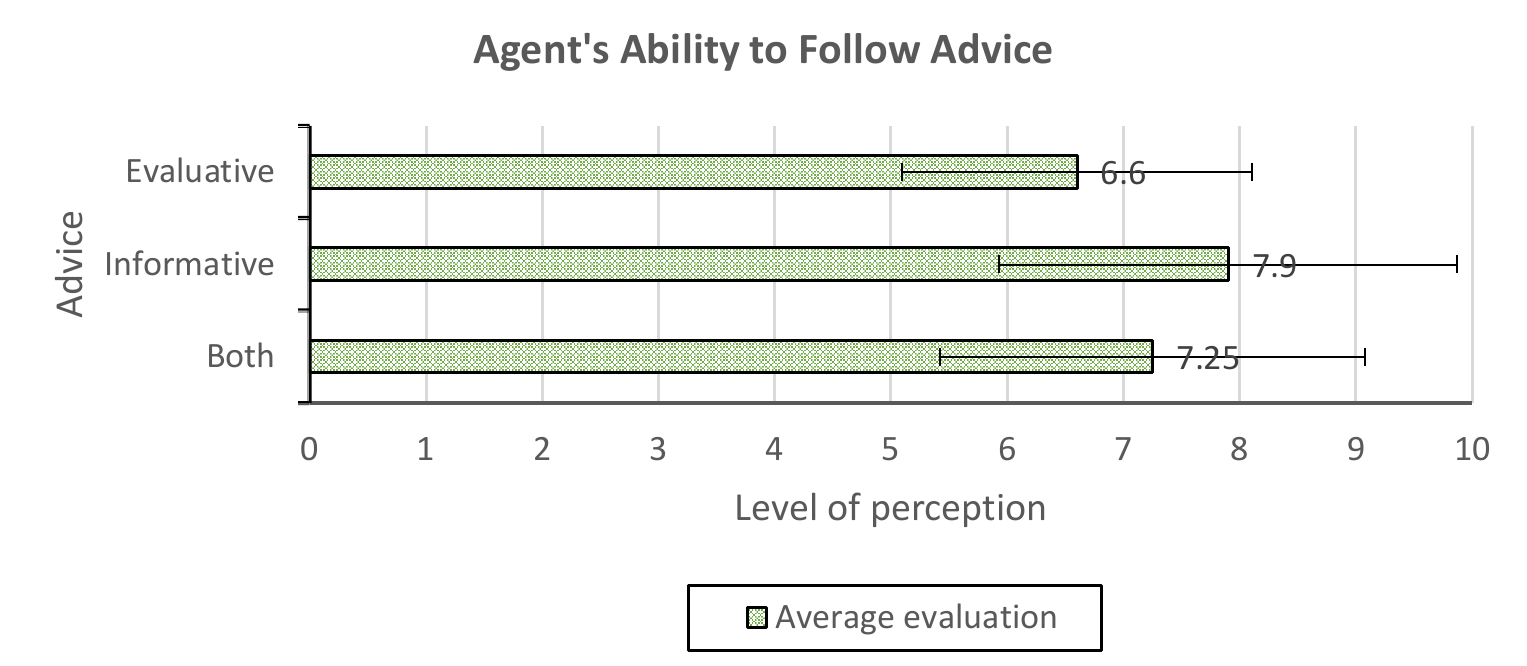}
    \caption{
    Average of participants' self-reported feeling of how well the agent followed the advice provided. 
    The participants score the agent's ability to follow advice using a scale from 0 (never), to 10 (always). 
    The informative group perceives the agent to better follow the provided advice. 
    }
    \label{fig:SelfFollowAdvice}
\end{figure}

Finally, participants were asked to rate how well they thought the agent followed their advice. 
On a scale from 0 (never), to 10 (always), participants scored the agent's ability to follow the given advice. 
The obtained results, summarised in Figure~\ref{fig:SelfFollowAdvice}, show that participants using informative advice perceived the agent as better able to follow advice when compared to participants using evaluative advice.

We have computed the Student's t-test to test the statistical difference between the self-reported results from the two groups.
We have used the threshold of 0.1146 for the p-values to consider the test statistically significant mainly due to the small number of participants.
Table~1 shows the obtained t-scores along with the p-values for the level of understanding of the environment, before and after the experiment, as well as the reported agent's ability to follow advice.
While the differences in the self-reported understanding of the environment previous to the experiments and the perceived agent's ability to follow the provided advice are statistically significant, the difference between the two groups in the self-reported understanding of the environment after the experiments is not significant, confirming the results reported in Figure~\ref{fig:SelfUnderstanding}.

\begin{table}
\caption{Student's t-test for comparison of self-reported results for evaluative and informative advisors.}
\begin{tabular}{p{4.1cm}ll} 
    \hline
    \textbf{Evaluation} & \textbf{t-score} & \textbf{p-value}\\
    \hline
    Understanding of the environment (before) & $t$=2.0608 & $p$=0.0541\\
    Understanding of the environment (after)  & $t$=0.3943 & $p$=0.6980\\
    Agent's ability to follow advice          & $t$=1.6584 & $p$=0.1146\\
    \hline
\end{tabular}
\label{table:SelfTstudent}
\end{table}

\subsection{Characteristics of the Provided Advice}

\begin{figure}[ht]
    \centering
    \includegraphics[width=0.85\linewidth]{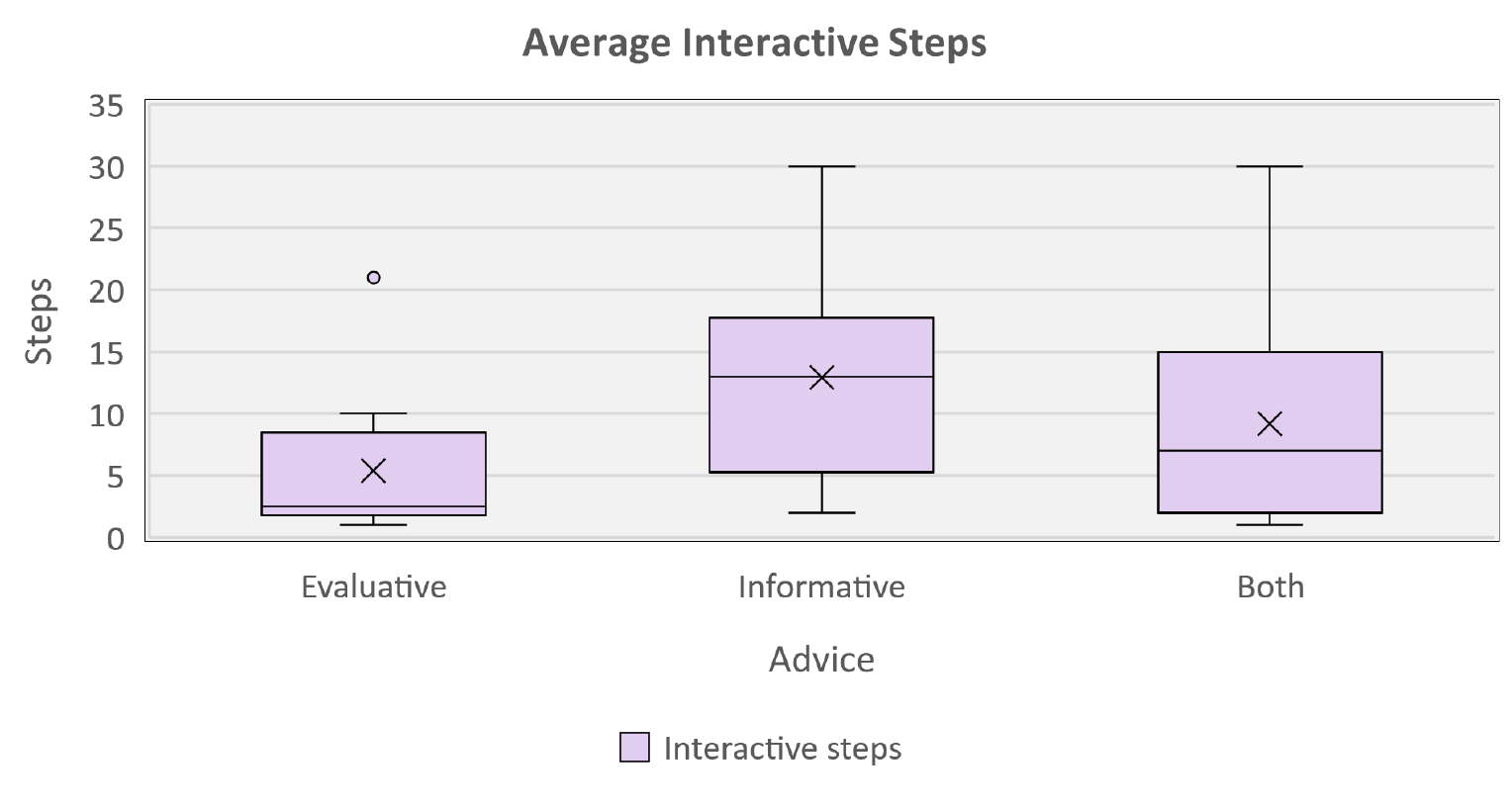}
    \caption{
    Number of steps that participants provided advice to the learner agent on the Mountain Car environment. 
    The amount of interactive steps is over two times for participants providing informative advice in comparison to evaluative advice. 
    }
    \label{fig:AverageInteractiveEpisodes}
\end{figure}

From the assistance provided to the agent, we kept a record of the number of interactive steps and the percentage relative to the total amount of steps.
Figure~\ref{fig:AverageInteractiveEpisodes} displays the number of steps that each set of participants interacted with the agent to provide assistance. 
In the boxplot, the cross marker represents the mean, dots are outliers, and the quartile calculation uses exclusive median. 
Overall, both groups provided advice in 9.15 steps on average, however, the data collected show a large variation in the engagement between the two types of advice. 
Participants providing informative advice assisted over twice as many steps than participants providing evaluative advice.

\begin{figure}[ht]
    \centering
    \includegraphics[width=1\linewidth]{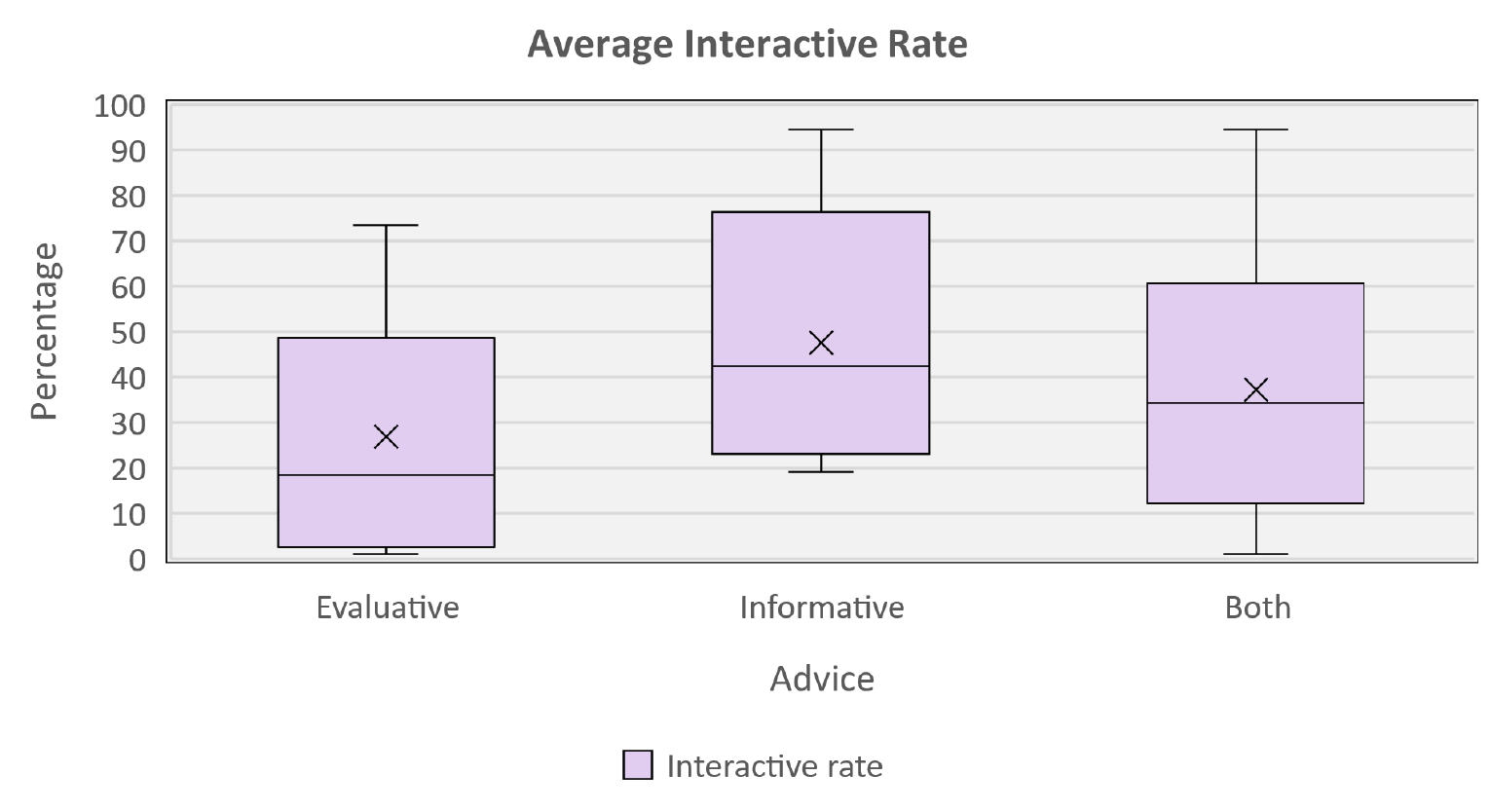}
    \caption{
    Percentage of steps that participants provided advice to the learner agent on the Mountain Car problem. 
    The percentage is computed as the ratio of interactions to interaction opportunities. 
    The informative advice rate is almost twice as high in comparison to evaluative advice.
    }
    \label{fig:AverageInteractiveRate}
\end{figure}

As demonstrated in previous work~\cite{griffith2013policy}, agents assisted by informative advice learn quicker than agents assisted by evaluative advice. 
The increase in learning speed results in fewer steps per episode for environments with a termination condition. 
This decrease in steps per episode for informative assisted agents gives fewer opportunities for the user to provide advice, as only one interaction may occur each step. 
As a result, the number of interactions per episode is not necessarily a suitable measure of engagement. 
Therefore, the number of steps in which interaction occurred relative to the total amount of steps is used to measure engagement. 
Figure~\ref{fig:AverageInteractiveRate} shows the interaction rate as a percentage for the two sets of participants. 
As before, the boxplot uses cross markers to represent the mean and exclusive median for quartile calculation.
The interaction percentage is the ratio of interactions to interaction opportunities. 
Using this measurement, the length of the episode is disregarded. 
The results show that participants using an informative advice delivery method interact almost twice as often as their evaluative counterparts. 
Despite the higher rate of interaction shown by participants using informative advice, both groups self-reported they were happy with their level of engagement with the agent, as shown in Figure~\ref{fig:SelfEngagement}. 

\begin{figure}[ht]
    \centering
    \includegraphics[width=1\linewidth]{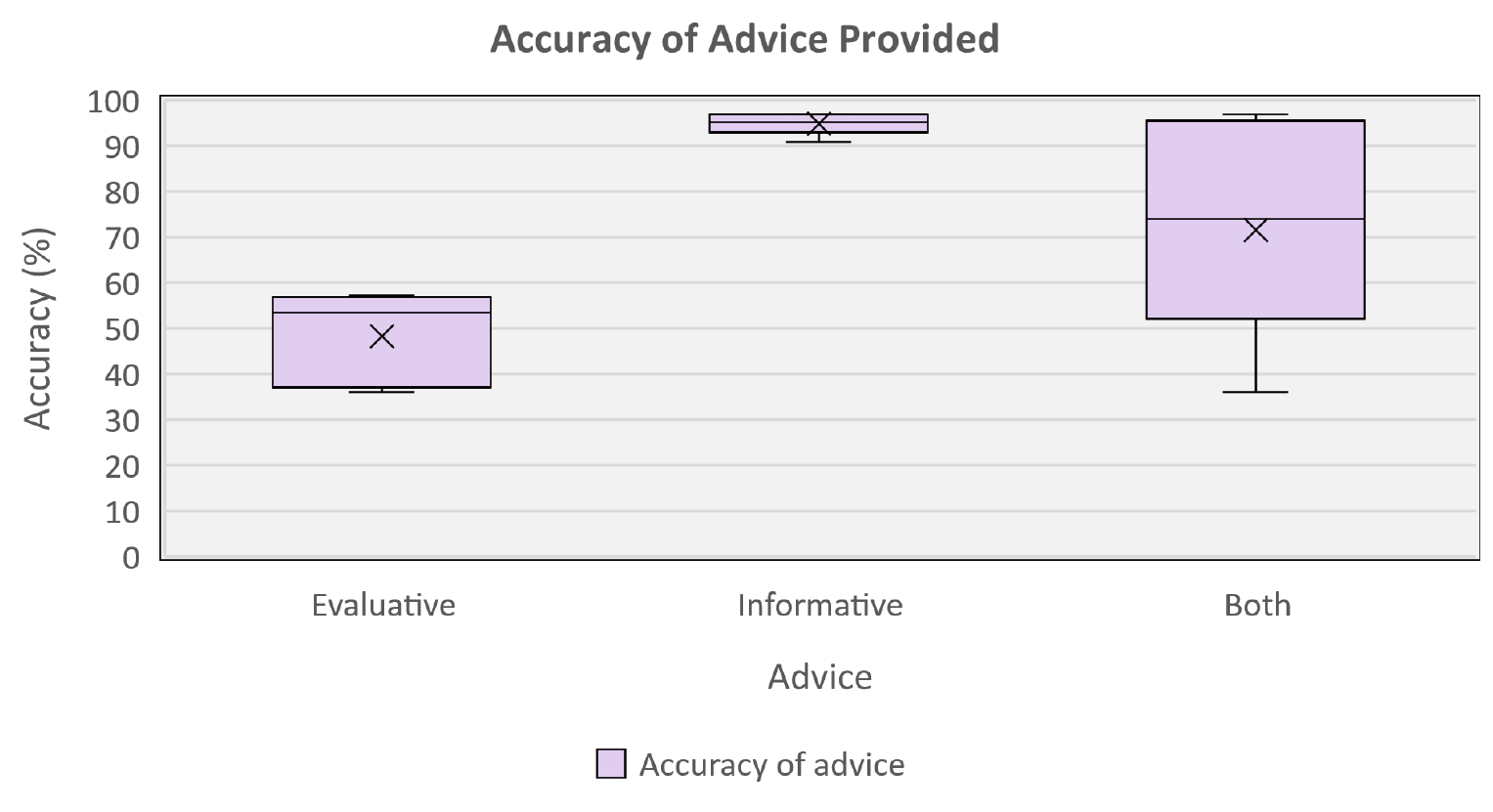}
    \caption{
    The percentage of interactions in which the advice provided was optimal for the state-action. 
    Participants providing informative advice were around two times more accurate and presented less variability in comparison to participants using evaluative advice.  
    }
    \label{fig:AccuracyAdviceProvided}
\end{figure}

While training the agent, the availability and accuracy of the provided assistance by the advisors were recorded.
Figure~\ref{fig:AccuracyAdviceProvided} displays the accuracy percentage of the advice provided by each of the groups of participants. 
Cross markers represent the mean and exclusive median is used for quartile calculation. 
An accurate interaction is one that provided the optimal advice for the agent in that state. 
Therefore, accuracy is a measurement of the number of correct interactions compared to the total interactions. 
Informative interactions are almost twice as accurate in comparison to evaluative interactions and also show much less variability. 
These results also reflect the self-reported level of advice accuracy shown in Figure~\ref{fig:SelfAccuracy}.

We have also computed the Student's t-test to test the statistical difference between the obtained results in terms of the advice provided from the two groups.
Table~2 shows the obtained t-scores along with the p-values for the average interactive steps, the average interactive rate, and the accuracy of the advice provided.
Although there exist statistically differences between the two groups for the average interactive steps and the average interactive rate, this is much clearer in the accuracy of the advice provided given the low p-value.

\begin{table}
\caption{Student's t-test for comparison of the provided advice from evaluative and informative advisors.}
\begin{tabular}{p{3.78cm}ll}
    \hline
    \textbf{Evaluation} & \textbf{t-score} & \textbf{p-value}\\
    \hline
    Average interactive steps      & $t$=2.2530  & $p$=0.0370\\
    Average interactive rate       & $t$=1.6828  & $p$=0.1097\\
    Accuracy of the advice provide & $t$=14.5772 & $p$=2.0778e-11\\
    \hline
\end{tabular}
\label{table:AdviceTstudent}
\end{table}

One hypothesis for the large difference in accuracy is latency. 
In this context, latency is the time it takes for the human to decide on the advice to provide, and then input it into the agent. 
It is possible that if the human is too late in providing advice, then the advice will inadvertently be provided to the state after the one intended. 
For the Mountain Car environment, a late interaction is more likely to remain accurate in the next state for informative advice than it is for evaluative advice. 
This is due to the layout of the state-space and the nature of untrained agents. 
The optimal action for a state in the Mountain Car environment is likely to be the same as its neighbouring states. 
This is due to the optimal behaviour being to accelerate in a single direction until velocity reaches 0. 
Therefore, a recommended action that is received in the state after the one intended is likely to be the correct action, regardless of latency. 
This does not apply to evaluative advice. 
The participants assisting the evaluative agent do not provide a recommended action, instead, they critique the agent's last choice. 
An untrained agent has a largely random action selection policy and is therefore not likely to choose the same action twice in a row. 
As the agent's chosen action may have changed by the time it receives advice from the user, the accuracy is more affected.

\begin{figure*}
    \centering
    \includegraphics[width=0.9\linewidth]{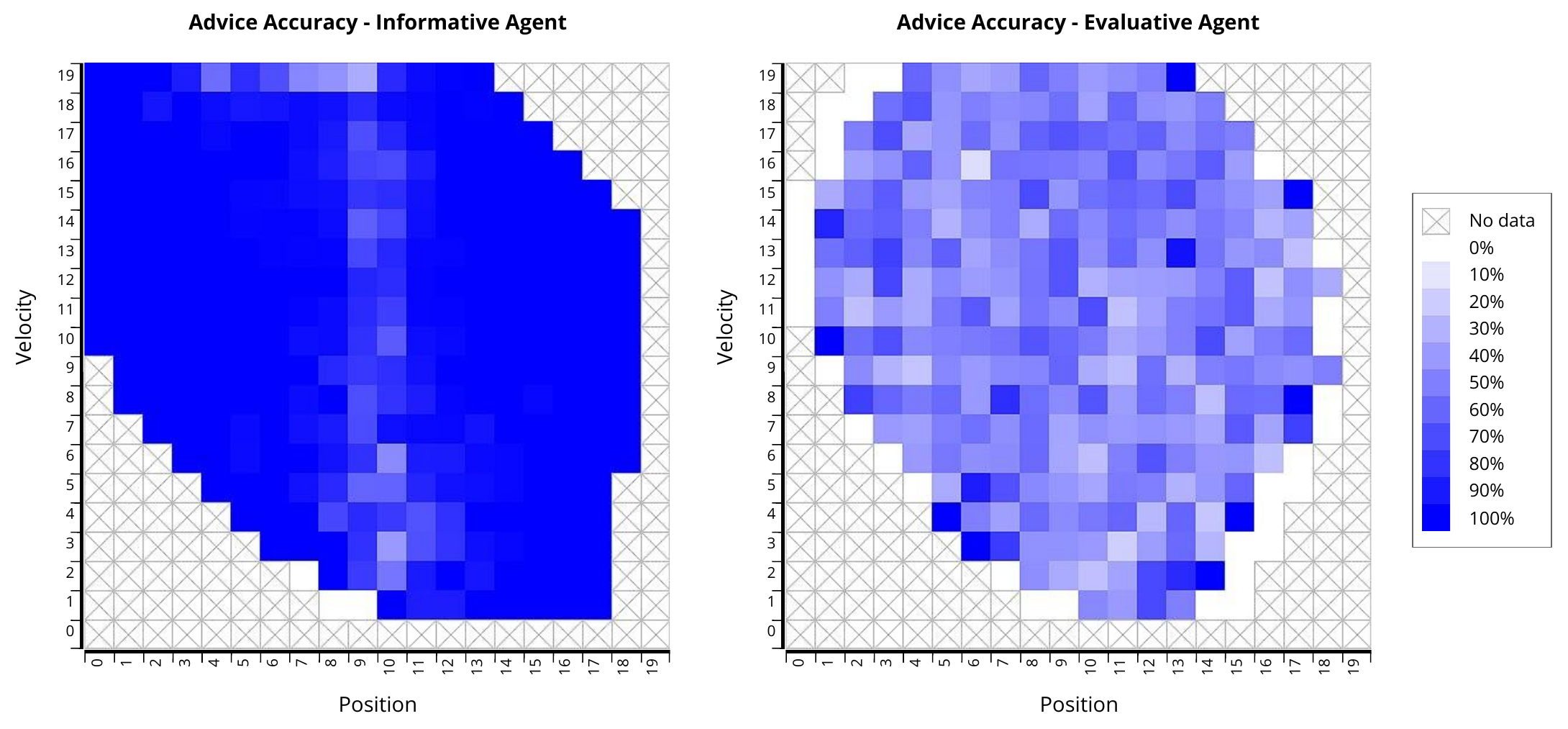}
    \caption{
    State-based accuracy of (a) informative and (b) evaluative participants for the Mountain Car environment.
    Informative advice is in general more accurate than evaluative advice, except in states in the middle of the environment, where the optimal action changes. 
    }
    \label{fig:AgentAccuracy}
\end{figure*}

This hypothesis is supported by the state breakdown of the advice accuracy. 
Figure~\ref{fig:AgentAccuracy} shows the accuracy of participants' advice for each state in the environment for (a) informative and (b) evaluative advice respectively. 
The darker the colour, the more accurate the advice supplied by the participants for that state. 
The comparison of the two heatmaps supports the earlier observations of the accuracy shown in Figure~\ref{fig:AccuracyAdviceProvided}; informative is much more accurate than evaluative advice. 
The informative advice method (Figure~\ref{fig:AgentAccuracy}a) shows that the states with the most inaccuracy are in the middle of the environment, where the optimal action changes. 
This inaccuracy is likely not due to poor participant knowledge, but rather providing delayed advice, after the agent has moved beyond the centre threshold. 

\begin{figure*}
    \centering
    \includegraphics[width=0.9\linewidth]{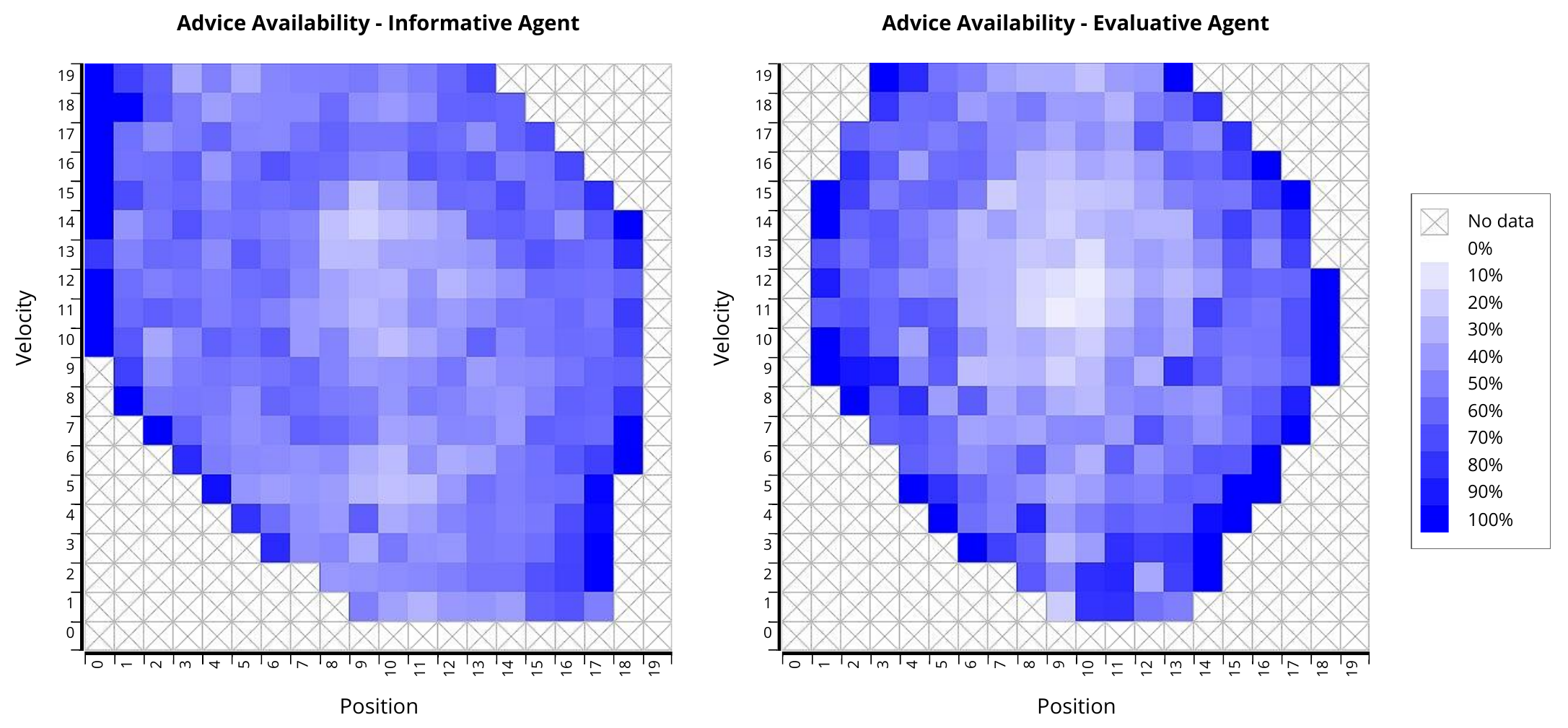}
    \caption{
    State-based availability of (a) informative and (b) evaluative participants for the Mountain Car environment.
    Agents following participants using informative advice achieved higher velocities in the environment, and as a consequence, more states were visited in comparison to the evaluative advice approach. 
    }
    \label{fig:AdviceAvailability}
\end{figure*}


The evaluative advice method (Figure~\ref{fig:AgentAccuracy}b) shows that accuracy differs wildly across the environment and does not have an obvious pattern. 
The poor result for accuracy of evaluative advice is likely due to the latency of advice delivery coupled with the lower probability that the advice will still be accurate to the following state compared to informative advice. 
Additionally, evaluative advice may have lower accuracy as it requires the human assessing each state-action pair. 
On the other hand, informative advice may require less time assessing each state, as the human may be following a set of rules for action recommendation, and that the next state is easier to predict compared to the agent's next action choice.

Figure~\ref{fig:AdviceAvailability} shows the availability of participants' advice for each state in the environment for (a) informative and (b) evaluative advice respectively. 
Availability in this context is a measure of how often the user provides advice in a state compared to the number of times the agent visited the state. 
The darker a state is on the heatmaps, the more often the user provides advice for that state. 
The agent that was assisted by informative advice (Figure~\ref{fig:AdviceAvailability}a) was able to achieve higher velocities in the environment, and as a result, visited more states in comparison to the evaluative advice method (Figure~\ref{fig:AdviceAvailability}b). 
One pattern that can be observed in the results is that the states on the edges show higher advice availability than those in the centre. 
These edge states are visited when the agent has learned a suitable behaviour, making the evaluation and recommendation of actions easier on the user, and increasing engagement. 
The edge states tend to be the last states the users provided advice, before voluntarily ending the experiment.

\begin{figure}[ht]
    \centering
    \includegraphics[width=1\linewidth]{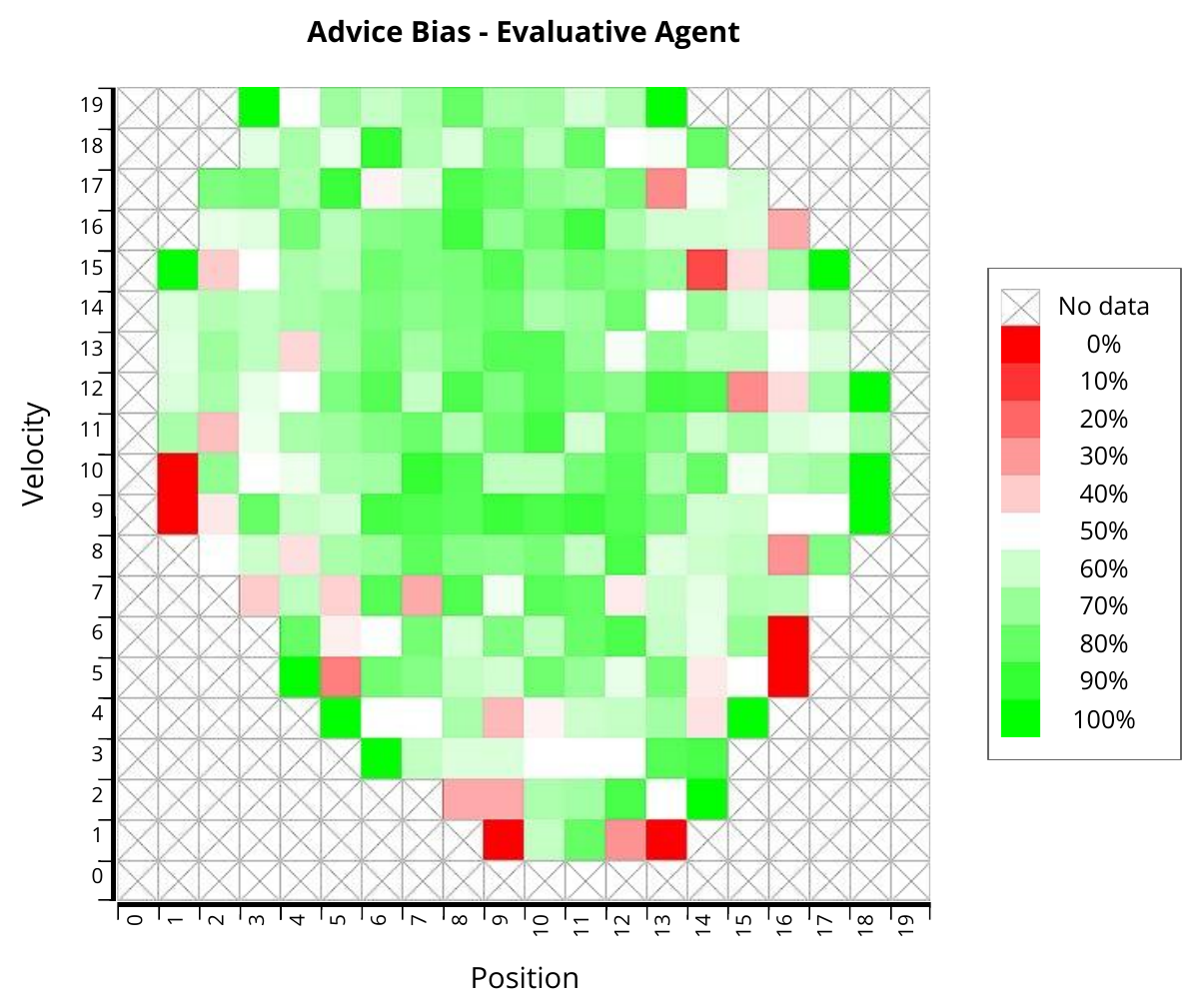}
    \caption{
    Reward bias of evaluative advice. 
    Above 50\% means that the advisor provided more positive evaluation than negative evaluation.
    }
    \label{fig:EvaluativeAdviceBias}
\end{figure}

Finally, we tested the presence of the reward bias of the participants providing evaluative advice as it has been reported in existing literature~\cite{amershi2014power}.
In this regard, a deviation from fifty percent indicates reward bias, i.e.,  above $50\%$ means that the advisor provided more positive evaluation than negative evaluation.
On average, participants provided $66.22\%$ of positive advice, with a minimum rate of $57.14\%$ and a maximum rate of $100.00\%$ of positive evaluations.
Figure~\ref{fig:EvaluativeAdviceBias} shows the reward bias of the participants providing evaluative advice. 
In general, participants provided much more positive advice, confirming prior findings that people are more likely to provide feedback on actions they view as correct than on incorrect actions.

\section{Conclusions}

The human trial performed in this work has investigated the engagement of human advice-givers when assisting interactive reinforcement learning agents. 
The assessment was performed using two methods for providing assistance, namely, evaluative and informative advice. 
Evaluative advice assesses the past performance of an agent, while informative advice supplements future decision-making. 
Previous work in the field has compared the performance of interactive reinforcement learning agents under the influence of each assistance method, finding that informative-assisted agents learn faster. 
However, to the best of our knowledge, studies on human engagement when providing advice using each assistance method have not been performed.

The results obtained from the human trial show that advice-givers providing informative advice outperformed those that used evaluative advice. 
Humans using an informative advice-giving method demonstrated more accurate advice, assisted the agent for longer, and provided advice more frequently. 
Additionally, informative advice-givers rated the ability of the agent to follow advice more highly, perceived their own advice to be of higher accuracy, and were similarly content with their engagement with the agent as the evaluative advice-giving participants. 

In this work, we have used a well-known environment and the advice was simply selecting the very next action. 
Therefore, this reduces considerably the subjectivity, however, there is a possibility of incorrect advice. 
The inconsistent advice problem has been previously studied~\cite{cruz2018improving, kessler2021interactive} and it has been shown that even a small amount of incorrect advice may be quite detrimental for the learning process.
This is especially relevant when the problem is more complex or the trainer is not familiar with the problem domain.
In such cases, human advice-givers need special additional preparation in order to avoid subjectivity. 
In this paper, advice is applied only to the current state and time-step, and so incorrect advice will not have a long-term impact. 
However, the problem becomes a major issue when general advice is given. 
For instance, in the persistent rule-based IntRL approach~\cite{bignold2021persistent} advice is retained by the agent and generalised over other similar states. 
To deal with such a  problem, that approach incorporates probabilistic policy reuse~\cite{fernandez2006probabilistic} which uses a decay factor in order to forget the given advice and trust more in the learned agent’s policy, unless new advice is given.

Future work will also consider the use of simulated users as a method of replicating the general behaviour of participants from this experiment.
Including simulated users would allow for faster experiments, keeping experimental conditions under control, and repeat the process as many time as needed. 
The findings from this study can be used to create simulated users which more closely reflect the behaviour of actual human advisers. 

\section*{Acknowledgments}
This work has been partially supported by the Australian Government Research Training Program (RTP) and the RTP Fee-Offset Scholarship through Federation University Australia.

\bibliographystyle{ieeetr}

\bibliography{references}

\newpage


\label{appendix:questionnaire}


\section*{Supplementary material}

# Questionnaire Form

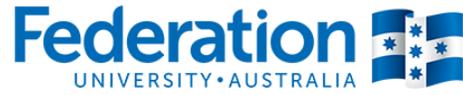

**Participant Code:** _____

No identifying information is collected. The participant code is used to match your questionnaire responses to your experiment responses. After completing, neither your name nor any identifying information will be kept. See the Plain Language Information Statement for more details.

**Have you participated in a machine learning study in the past?**

_____

_____

**On a scale of 0 – 10, how would you rate your level of knowledge about the Mountain Car experiment?**

       0 (Nothing)    1    2    3    4    5    6    7    8    9    10 (Expert)

**<span style="color:red">Stop the questionnaire now and perform the experiment. After completing the experiment, turn over the page and complete the questions.</span>**

CRICOS Provider No: 00103D     Benchmarking Human Advice Giving Study – Questionnaire Form 2017     **0** | P a g e

22

# Questionnaire Form

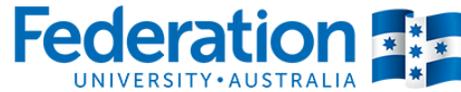

**Do not complete this side until you have completed the experiment.**

**Now that you have completed the experiment, on a scale of 1 – 10, how would you rate your level of knowledge about the Mountain Car experiment?**

      0 (Nothing)   1   2   3   4   5   6   7   8   9   10 (Expert)

**How do you feel about the level of engagement you had with the agent?**

A. I could have spent more time interacting with the agent.
B. I'm happy with how much time I interacted with the agent.
C. I spent too much time interacting with the agent.

**How accurate do you feel your advice was to the agent?**

A. Always Incorrect
B. Mostly Incorrect
C. Sometimes Incorrect
D. Sometimes Correct
E. Mostly Correct
F. Always Correct.

**How well do you think the agent followed your advice?**

      0 (Never)   1   2   3   4   5   6   7   8   9   10 (Always)

**Is there any other information you want to provide?**



\end{document}